# 基于双目视觉的无人车高精度目标定位系统

贺先祺,李子睿,尹旭峰,龚建伟,龚乘

(北京理工大学 机械与车辆学院,北京 100081)

## 摘要

无人车在工作过程中,常常需要对目标进行高精度的定位。在无人运料车间,无人车需要对工件进行高精度的位姿估计以准确抓取工件。在此背景下,本文提出一种基于双目视觉的高精度无人车目标定位系统,该系统利用基于区域的立体匹配算法获取视差图,并利用 RANSAC 算法提取位置和姿态特征,实现了空间中六自由度圆柱体工件的位置与姿态估计。为了验证系统效果,本文采集了圆柱体在不同位姿下输出结果的精度和计算时间,实验数据显示该系统的位置精度为 0.61~1.17mm,姿态精度为 1.95~5.13°,能够实现较好的高精度定位效果。

**关键字** 双目视觉;车载传感器;无人车;基于区域的匹配;高精度视觉定位

## 一、 引言

无人车在生产中的应用可以极大提高生产效率。在无人运料车间,无人车可包办"危脏难"任务,提高生产安全性和生产品质。而无人车作业时,需要进行毫米级的操作,这对传感器的识别精度提出了更高要求。因此,研究并开发一种高精度的目标定位系统可提高生产效率和准确性,具有现实意义。

在无人车高精度定位领域有多种传感器方案,例如声呐、激光雷达、毫米波雷达、红外线传感器等。随着机器视觉技术与相关工艺的发展,基于视觉的方法正在成为研究热点。相比其他方法,机器视觉以其信息量大、定位精度高、通用性强、无接触等优势逐渐成为主流的方案。机器视觉按摄像头从少到多依次分为单目视觉、双目视觉、多目视觉。由于单目视觉无法提取目标点深度,多目视觉部署复杂,而双目视觉在实现深度提取的同时计算量适中,因此本文选用双目视觉的部署方案。

对于双目视觉的高精度目标定位问题,研究者们在这一领域进行了诸多工作以提升算法的性能。在此论文[1]中,Ekvall, S.等人设计了一种工件的位姿识别方法用于抓握工件。这种方法基于颜色直方图和几何建模,并应用了经典学习框架制定决策;Aldoma, A.等人通过比较算法内 CAD 模型和实际工件的差异进行工件的识别与匹配[2]。文章提出了一种基于 CAD 模型的三维重建方法,这种方法首先获取工件的图像,然后根据获得的图像建模,通过比较建模得到的点云与标准 STL 的差异校准三维重建的点云。

此外,在文章[3]中,Morency, L.等人使用了基于 3D 视图的本征空间。在文章[4]中,Rusu, R.B.等人应用了特征直方图。在文章[5]中,Azad, P.等人针对无颜色特征工件的识别进行了基础研究。在文章[6]中,Changhyun, C.等人为抓取工件设计了一种基于角点特征的识别方法。在文章[7]中,Mittrapiyanumic, P.等人提出了一种主动外表识别的方案。在文章[8]中,Payet, N.使用物体的等高线图作为数据输入,提出了一种具有视角不变性的工件识别与位姿估计算法。在文章[9]中,Drost, B.等人设计了一种在三维点云中识别工件的方案。这种方案设计了一种快速投票系统,这一投票系统基于旋转不变的匹配点对,可用于识别点云中局部的工件模型。

无人车识别的目标常常是无颜色特征无纹理的物体，且包含圆柱体特征。因此，本文提出针对无颜色无纹理的圆柱体工件的高精度目标定位系统。该系统能对工件进行三维重建，并从重建的点云中提取几何要素，计算工件的位置和姿态特征。

首先，本文分析了摄像头成像模型和视差测距原理，选取平行双目视觉模型搭建视觉定位系统；其次研究了双目相机的标定方法，比较后选取张正友标定法作为标定方法，并在之后提出了一种结合传统标定法和张正友标定法的改进方法；接着研究并比较了不同立体匹配算法，优选基于区域的立体匹配算法以及 SAD 灰度匹配相似性度量函数进行立体匹配；然后研究并比较不同的位姿估算方法，优选分步提取的工作流程，并选择 RANSAC 提取特征；最后设计实验验证了双目定位系统的有效性，分析了影响精度的因素。

因此，本文的主要工作双目视觉定位分为三个阶段，如图 1 所示：

（1）立体匹配。利用基于区域的立体匹配方法计算视差图。

（2）三维重建。根据提取的匹配点计算深度，重建目标物体的三维坐标。

（3）从三维重建的点云中提取信息。

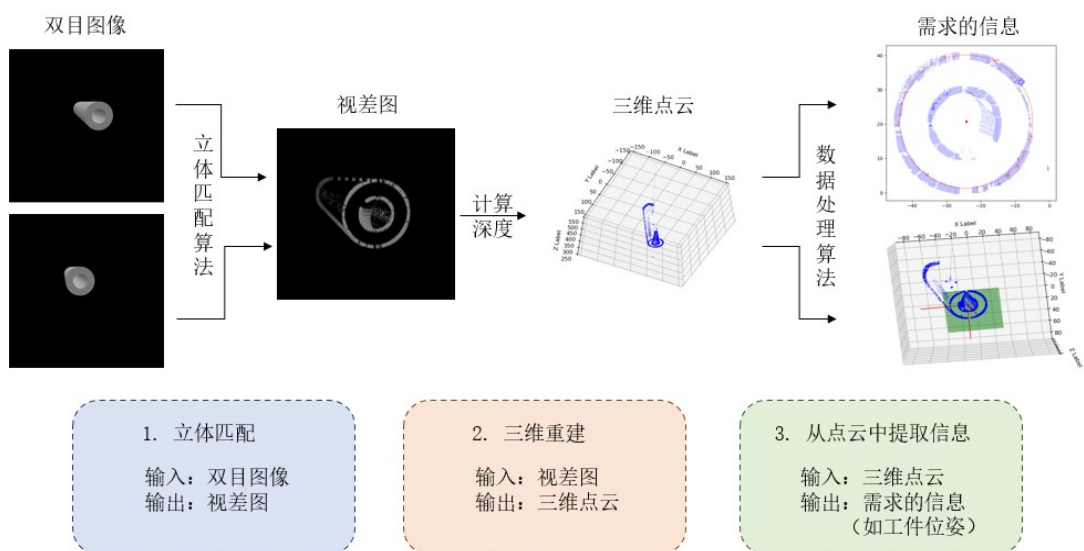

图 1 双目视觉工作流程

## 二、 目标识别

### 2.1 双目视觉模型

摄像机采用线性成像模型。为了定义摄像机的成像过程，需要定义以下几个坐标系：

（1）世界坐标系 $(X_W, Y_W, Z_W)$，客观世界的绝对坐标系。

（2）摄像机坐标系 $(X_C, Y_C, Z_C)$，以相机光心为坐标原点，通常沿光轴方向为 $Z$ 轴，且与成像平面垂直，光轴与图像平面交点 $O_P = (u_0, v_0)$ 为图像主点，$O_C O_P$ 为摄像机焦距 $f$。

（3）图像物理坐标系 $(X_P, O_P, Y_P)$，以光轴和像平面的交点为原点，即图像主点 $O_P = (u_0, v_0)$，其中 $X$ 轴 $Y$ 轴分别平行于摄像机坐标系的 $X$，$Y$ 轴，以毫米为单位，是 2D 平面直角坐标系。

（4）图像像素坐标系$(U,O,V)$，即计算机帧存坐标系，是图像上的平面直角坐标系，一般以像素为单位，图像左上角为其原点，$U$，$V$轴分别平行于图像物理坐标系的$X$，$Y$轴。像素坐标$(u,v)$分别表示该像素在图像上的列数和行数，即$(u,v)$为以像素为单位的图像坐标[10]。

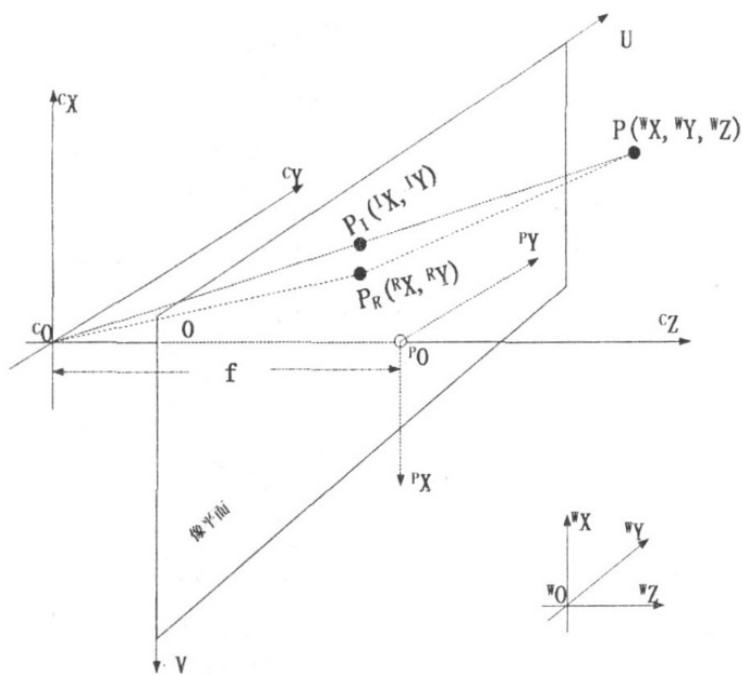

图 2 摄像机线性成像模型

四个坐标系之间的变换关系如下：

（1）世界坐标系与摄像机坐标系之间的变换。

世界坐标系与摄像机坐标系均为笛卡尔坐标系，为方便起见，定义两坐标系的坐标尺度一致，坐标变换即可简化为刚性变换问题。两坐标系的变换关系可用平移矩阵$T$和单位正交矩阵$R$表示：

$$\begin{bmatrix} X_C \\ Y_C \\ Z_C \end{bmatrix} = R \begin{bmatrix} X_W \\ Y_W \\ Z_W \end{bmatrix} + T = \begin{bmatrix} r_{11} & r_{12} & r_{13} \\ r_{21} & r_{22} & r_{23} \\ r_{31} & r_{32} & r_{33} \end{bmatrix} \begin{bmatrix} X_W \\ Y_W \\ Z_W \end{bmatrix} + \begin{bmatrix} t_x \\ t_y \\ t_z \end{bmatrix} \quad (1)$$

其中，矩阵$R$为正交旋转矩阵，$T$为世界坐标系的原点在摄像机坐标系中的表示。将式(1)齐次表示为：

$$\begin{bmatrix} X_C \\ Y_C \\ Z_C \\ 1 \end{bmatrix} = \begin{bmatrix} R & T \\ 0^T & 1 \end{bmatrix} \begin{bmatrix} X_W \\ Y_W \\ Z_W \\ 1 \end{bmatrix} \quad (2)$$

（2）图像坐标系与摄像机坐标系之间的变换。

摄像机坐标系中特征点$P$在图像物理坐标系中的像点$P_1$坐标如图 2，可表示为：

$X_I = X_C \cdot f / Z_C$ ，$Y_I = Y_C f / Z_C$。齐次坐标表示为：

$$Z_C \begin{bmatrix} X_I \\ Y_I \\ 1 \end{bmatrix} = \begin{bmatrix} f & 0 & 0 & 0 \\ 0 & f & 0 & 0 \\ 0 & 0 & 1 & 0 \end{bmatrix} \begin{bmatrix} X_C \\ Y_C \\ Z_C \\ 1 \end{bmatrix} \tag{3}$$

（3）图像坐标系与像素坐标系之间的变换。

不管是 CMOS 还是 CCD 相机，其传感器均由方形感光元件紧密排列而成。图像的像素坐标系取在图像的中心，也就是图像的主点。但由于制造工艺缺陷主点往往会偏移图像中心。假设主点的坐标为 $(u_0, v_0)$，像素坐标与图像坐标的转换公式为：

$$\begin{cases} u - u_0 = X_I / d_x = s_x X_I \\ v - v_0 = Y_I / d_y = s_y Y_I \end{cases} \tag{4}$$

齐次坐标表示为：

$$\begin{bmatrix} u \\ v \\ 1 \end{bmatrix} = \begin{bmatrix} s_x & 0 & u_0 \\ 0 & s_y & v_0 \\ 0 & 0 & 1 \end{bmatrix} \begin{bmatrix} X_I \\ Y_I \\ 1 \end{bmatrix} \tag{5}$$

其中 $u_0$，$v_0$ 是主点坐标，$d_x$，$d_y$ 分别为一个像素在 $P_X$，$P_Y$ 方向上的物理尺寸，$s_x = 1/d_x$，$s_y = 1/d_y$ 表示单位长度上的像素个数。实际情况下由于加工和安装误差，摄像头的光轴往往并不垂直于成像平面，图像像素会产生倾斜，实际坐标修正为：

$$\begin{cases} u = u' - \dfrac{v}{\tan \alpha} \\ v = \dfrac{v'}{\sin \alpha} \end{cases} \tag{6}$$

修正倾斜后式(5)变为：

$$\begin{bmatrix} u \\ v \\ 1 \end{bmatrix} = \begin{bmatrix} s_x & \tan \alpha & u_0 \\ 0 & s_y & v_0 \\ 0 & 0 & 1 \end{bmatrix} \begin{bmatrix} X_I \\ Y_I \\ 1 \end{bmatrix} \tag{7}$$

综合上述推导，令 $s = \tan \alpha$，对于世界坐标系中的空间点 $P$ 与像素点 $P_I$ 之间的坐标变换关系为：

$$Z_C \begin{bmatrix} u \\ v \\ 1 \end{bmatrix} = \begin{bmatrix} f_x & s & u_0 & 0 \\ 0 & f_y & v_0 & 0 \\ 0 & 0 & 1 & 0 \end{bmatrix} \begin{bmatrix} R & T \\ 0^T & 1 \end{bmatrix} \begin{bmatrix} X_W \\ Y_W \\ Z_W \\ 1 \end{bmatrix} = M_1 M_2 X = MX \tag{8}$$

其中 $f_x = f/d_x$，$f_y = f/d_y$；$M$ 为投影矩阵；$M_1$ 完全由摄像机的内部结构决定，故被称为内参数矩阵[11]或重投影矩阵。

## 2.2 摄像机标定

拍摄的工件二维图像和三维空间中的世界坐标有确定的转换关系。如果得到摄像机的内参矩阵、外参矩阵、畸变系数，那么就能根据内外参矩阵逆推出像素点在世界坐标系中的坐标，从而重建三维点云，此过程即为双目视觉系统的标定。

标定方法可分为三种：传统标定法、主动视觉标定法、自标定法三种。本文采用结合了传统标定法与自标定法的张正友法进行标定，并进行了改进。原始的张正友法并不能达到精度要求，标定后存在 4~5mm 的标定误差，因此在张正友法上进行改进。

改进方法先将误差集中在 z 轴（深度方向）：采集图像时，保持标定板的 z 坐标不变，然后变换 x、y 坐标使棋盘格均匀充满摄像机的视场。该种标定方法得到的结果在 xy 方向上标定误差小于 0.5mm，有很好的精度；在 z 轴方向上有 4~5mm 的误差，这表明误差已经集中在了 z 轴上。

接着，用传统标定法修正 z 轴误差。依据世界坐标系到相机坐标系的转换关系，z 轴标定误差主要与焦距、光轴倾斜度、外参矩阵、切向畸变与径向畸变四个因素有关。上一步中我们已经用张正友标定法进行第一次标定，外参矩阵、径向畸变与切向畸变已测量完毕，而仿真环境中的相机为理想相机，光轴倾斜度为 0，因此仅需考虑焦距偏差对 z 轴方向精度的影响，即 $Z = f - \dfrac{f \cdot D_x}{x_2 - x_1} = f - \dfrac{f \cdot D_y}{y_2 - y_1}$。本文的双目视觉系统采用平行双目视觉模型，匹配点在 $y$ 轴方向的视差为 0，因此深度仅和 $x$ 轴方向的视差相关。令视差 $d = x_2 - x_1$，深度 $Z = f(1 - \dfrac{D}{d})$，当视差 $d$ 保持不变时，深度 $Z$ 与焦距 $f$ 成线性关系。假设焦距真值为 $f_0$，深度真值为 $Z_0$，测得的深度为 $Z$，张正友标定法解算得的焦距为 $f$，可得 $f_0 = f \cdot \dfrac{Z_0}{Z}$。经测试最终在 x、y 轴方向上标定误差小于 0.5mm，在 z 轴方向上标定误差小于 0.3mm，能够满足高精度双目视觉系统的要求。

## 2.3 立体匹配

双目视觉系统通过视差图计算深度，而视差图来自立体匹配算法。一般来说，立体匹配方法分为基于特征的立体匹配算法和基于区域的立体匹配算法两种。基于特征的匹配算法通过如角点、轮廓边等特征识别像素对，现有的算法如 Harris 角点、SIFT、SURF、FAST 等算法可根据物体表面的纹理和光线变化匹配左右眼图像中的特征点。基于区域的特征匹配算法则是针对图像 A 中的某个像素点，通过一个 m*n 大小的滑动窗口，在图像 B 中以卷积的形式扫描，选取相似率最高的区域作为匹配点。这种算法与基于特征的匹配算法最大的区别是基于特征方法提取的匹配点是独立的角点或边缘，呈稀疏状；而基于区域方法提取的特征点经常有连续的线段、平面，呈稠密状。

SIFT 算法是一种典型的基于特征的立体匹配算法。当表面富有纹理时，SIFT 效果良好；而表面缺乏纹理和颜色时，立体匹配效果很差，如图 3 所示。

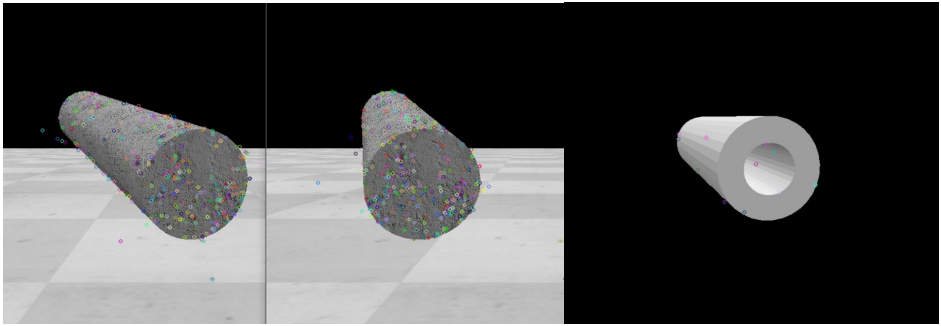

图 3 有纹理（左）和无纹理（右）表面上的 SIFT 算法

基于区域的匹配算法对无纹理表面则有较好效果。基于区域的匹配算法一般流程分为两步：（1）滑动窗口卷积扫描；（2）计算灰度匹配相似性度量函数。不同的基于区域的匹配算法，其本质区别就是相似性度量函数的不同，有 SAD、SSD、STAD、NCC、ZNCC 等。综合考虑时间成本与精度，采用 SAD 相似性度量函数进行立体匹配，表达式如下，其中 d 为像素灰度视差：

$$\text{SAD} = \frac{1}{(2m+1)(2n+1)} \sum_{i=-m}^{m} \sum_{j=-n}^{n} \left| g_1(u+i, v+j) - g_2(u+i, v+j+d) \right| \quad (9)$$

使用 SAD 算法提取的视差图如图 4 所示。

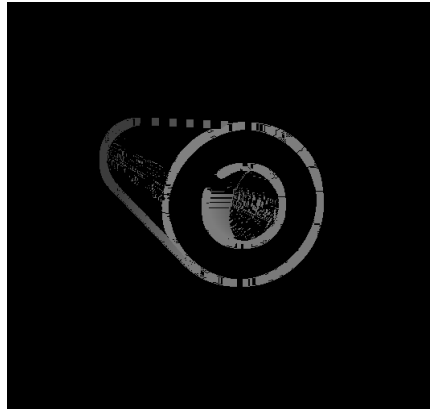

图 4 使用 SAD 算法提取的视差图

## 2.4 三维重建

三维重建是指利用二维投影或影像回复物体三维信息和形状的数学过程和计算机技术。本文所述的双目视觉定位系统采用平行双目视觉布置，假设世界坐标系中有一物体坐标 $P(X_w, Y_w, Z_w)$，并且在像素坐标系中获得其左摄像机坐标系中坐标为 $(u_l, v_l)$，在右摄像机坐标系中坐标为 $(u_r, v_r)$。根据式(8)，已标定测得左右两摄像机的内外参数，假设左相机的投影矩阵为 $M_l$，右相机的投影矩阵为 $M_r$：

$$M_l = \begin{bmatrix} l_{11} & l_{12} & l_{13} & l_{14} \\ l_{21} & l_{22} & l_{23} & l_{24} \\ l_{31} & l_{32} & l_{33} & l_{34} \end{bmatrix} \quad M_r = \begin{bmatrix} r_{11} & r_{12} & r_{13} & r_{14} \\ r_{21} & r_{22} & r_{23} & r_{24} \\ r_{31} & r_{32} & r_{33} & r_{34} \end{bmatrix} \quad (10)$$

将标定得到的 $M_l$ 和 $M_r$ 式(8)，得到：

$$Z_C^l \begin{bmatrix} u_l \\ v_l \\ 1 \end{bmatrix} = \begin{bmatrix} l_{11} & l_{12} & l_{13} & l_{14} \\ l_{21} & l_{22} & l_{23} & l_{24} \\ l_{31} & l_{32} & l_{33} & l_{34} \end{bmatrix} \begin{bmatrix} X_W \\ Y_W \\ Z_W \\ 1 \end{bmatrix} \qquad (11)$$

$$Z_C^r \begin{bmatrix} u_r \\ v_r \\ 1 \end{bmatrix} = \begin{bmatrix} r_{11} & r_{12} & r_{13} & r_{14} \\ r_{21} & r_{22} & r_{23} & r_{24} \\ r_{31} & r_{32} & r_{33} & r_{34} \end{bmatrix} \begin{bmatrix} X_W \\ Y_W \\ Z_W \\ 1 \end{bmatrix} \qquad (12)$$

联立(11)和(12),得到:

$$\begin{cases} (u_l l_{31} - l_{11})X_W + (u_l l_{32} - l_{12})Y_W + (u_l l_{33} - l_{13})Z_W = l_{14} - u_l l_{34} \\ (v_l l_{31} - l_{21})X_W + (v_l l_{32} - l_{22})Y_W + (v_l l_{33} - l_{23})Z_W = l_{24} - v_l l_{34} \\ (u_r r_{31} - r_{11})X_W + (u_r r_{32} - r_{12})Y_W + (u_r r_{33} - r_{13})Z_W = r_{14} - u_r r_{34} \\ (v_r r_{31} - r_{21})X_W + (v_r r_{32} - r_{22})Y_W + (v_r r_{33} - r_{23})Z_W = r_{24} - v_r r_{34} \end{cases} \qquad (13)$$

求解式(13)即可得到物体在世界坐标系的坐标 $P(X_W, Y_W, Z_W)$,从而得到达到三维重建的目的。将式(13)转写为矩阵:

$$\begin{bmatrix} u_l l_{31} - l_{11} & u_l l_{32} - l_{12} & u_l l_{33} - l_{13} \\ v_l l_{31} - l_{21} & v_l l_{32} - l_{22} & v_l l_{33} - l_{23} \\ u_r l_{31} - r_{11} & u_r r_{32} - r_{12} & u_r r_{33} - r_{13} \\ v_r l_{31} - r_{21} & v_r r_{32} - r_{22} & v_r r_{33} - r_{23} \end{bmatrix} \begin{bmatrix} X_W \\ Y_W \\ Z_W \end{bmatrix} = \begin{bmatrix} l_{14} - u_l l_{34} \\ l_{24} - v_l l_{34} \\ r_{14} - u_r r_{34} \\ r_{24} - v_r r_{34} \end{bmatrix} \qquad (14)$$

为便于表示,令:

$$C = \begin{bmatrix} u_l l_{31} - l_{11} & u_l l_{32} - l_{12} & u_l l_{33} - l_{13} \\ v_l l_{31} - l_{21} & v_l l_{32} - l_{22} & v_l l_{33} - l_{23} \\ u_r l_{31} - r_{11} & u_r r_{32} - r_{12} & u_r r_{33} - r_{13} \\ v_r l_{31} - r_{21} & v_r r_{32} - r_{22} & v_r r_{33} - r_{23} \end{bmatrix} \cdot W = \begin{bmatrix} X_W \\ Y_W \\ Z_W \end{bmatrix} \cdot D = \begin{bmatrix} l_{14} - u_l l_{34} \\ l_{24} - v_l l_{34} \\ r_{14} - u_r r_{34} \\ r_{24} - v_r r_{34} \end{bmatrix} \qquad (15)$$

式(14)转写为:

$$C \cdot W = D \qquad (16)$$

根据最小二乘法得到:

$$W = (C^T C)^{-1} C^T D \qquad (17)$$

求解式(17)即可计算出像素坐标系中匹配点在世界坐标系中的坐标,重建结果如图 5 所示。

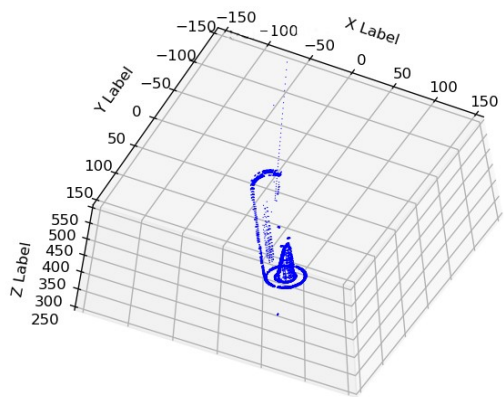

图 5 三维重建结果

## 2.5 位姿估计

对于圆柱外形的工件，姿态特征可用轴线向量、母线向量、端平面法向量等几何要素表示，位置特征可用轴线中心、母线中心、端平面中心、等几何要素表示。

第一种方案是识别圆柱外廓。圆柱外廓的母线向量可以表征姿态，母线中心、母线端点可以表征位置。但是，这一方法仅在圆柱表面有光泽时才能使用，当圆柱表面粗糙时，将不会呈现反射效果，摄像机捕捉不到母线；同时在仿真环境下，光照系统简单，圆柱表面没有复杂的阴影过渡，因此这一方法不可行。

第二种方案是识别圆柱端面的孔。孔的定位比检测圆柱外形简单，同时孔是工件上最突出的特征，即使光线条件差、工件表面无特征也可以清楚地看到。此外，识别孔的算法鲁棒性较强，在不同的拍摄角度下均可以看到。

综上所述，采用第二种识别孔的方法作为位姿估计方法。

识别圆孔的第一种办法是检测圆孔投影在成像平面上形成的椭圆。通过检测椭圆的长轴、短轴以及椭圆的倾斜度可出逆向推算出圆的空间姿态，进而计算套筒的旋转向量。然而由于通过椭圆的透视变换，逆推圆孔位姿的精度不高。

第二种方案将检测圆孔分成两步：先拟合端面所在的空间平面，再将端面点投影在平面上。这种方案提出的在二维平面上拟合平面圆算法成熟，通用性也比拟合椭圆好。拟合圆相较拟合椭圆有下列优势：第一，圆的拟合精度比椭圆的拟合精度高；第二，从椭圆逆推得圆的状态对拟合误差非常敏感，而在平面上拟合圆精度更高效果更稳定，其稳定性和准确性均满足高精度定位要求。[12]因此采用第二种分步拟合的工作流程。

系统提取位姿的步骤如下：

（1）数据预处理，提取端面点云

（2）RANSAC 算法拟合端面

（3）将点云投影到三维平面上，再将三维平面连同平面上的点转为二维平面

（4）二维平面上 RANSAC 拟合圆，二维圆心逆推回世界坐标系

第一步拟合圆所在的平面，采用 RANSAC（Random Sample Consensus）算法进行拟合。拟合平面前为提高拟合的准确性和快速性，对点云进行预处理。

视差测距形成的点云，其分布特征是有迹可循的。点云根本上来自于立体匹配算法对特征的识别，影响识别效果的因素有两个：(1) 特征到成像平面的距离；(2) 特征在立体匹配

算法中的显著度。一般来说，特征离成像平面越近，点云越大越密集；同时，特征与周围像素的区分度越大，点云越大越密集。在成像时，端面圆是工件所有部位中离成像平面最近的部分，同时端面圆还上有显著的内外圆轮廓线，可以推导得端面圆的点云是最大最密集的。基于这一特点，本文提出一种基于点云密度的数据预处理方法。

对于重建的三维点云，绝大多数的点都分布于端面上：整个点云合计 12000~13000 个点，其中分布在端面的点有 6000~7000 个，可以通取出聚集密度较大的点云，将之标记为端面点，然后舍掉非端面点，使平面拟合变得更容易。成功提取的端面点云如图 6 所示。

该数据预处理方法减轻了算法计算量，运算时间减少了 21.4%；同时减少了无关点投影在平面上的数目，提高后续定心精度。

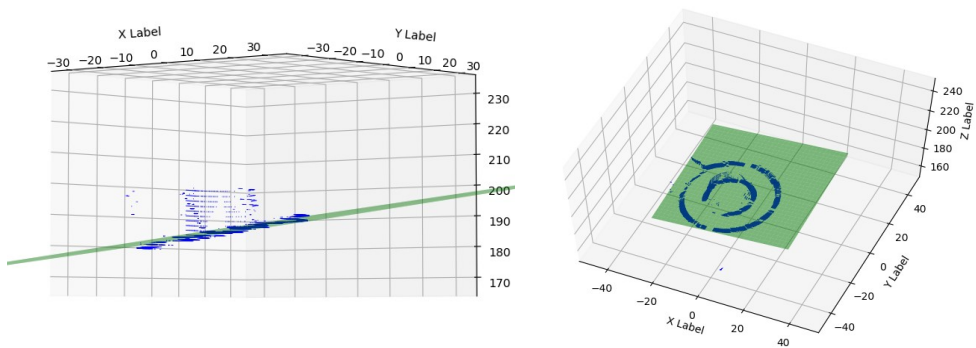

图 6 提取得到的端平面点云

拟合得到平面后，下一步将在平面上拟合圆。首先将所有点投影在平面上，由于圆柱的几何特性，当三维重建正确无重大误差时，点云全部投影在端面圆内。

给定的平面方程 $ax+by+cz+d = 0$，空间中任一点 $V_i = (x_i, y_i, z_i)$，令点在平面上的投影为 $V_i' = (x, y, z)$，直线 $V_iV_i'$ 与平面法向量 $\vec{N} = (a, b, c)$ 平行，直线 $V_iV_i'$ 的参数方程为：

$$\begin{cases} x = x_i - At \\ y = y_i - Bt \\ z = z_i - Ct \end{cases} \quad (18)$$

将 $x, y, z$ 代入(18)，求出 t，再将其代入(18)求出投影点 $V_i' = (x, y, z)$ 即可计算出投影点坐标。

得到二维平面后，利用仿射变换将三维空间中的平面转化为二维平面。这是一个坐标系变换问题。在平面 $ax+by+cz+d = 0$ 上任取两点 A 和 B，得到向量 $\vec{AB}$，又平面的法向量为 $\vec{N} = (a, b, c)$，归一化得到单位向量 $u\vec{AB}$ 和 $u\vec{N}$，由这两个单位向量建立第三个基本向量 $u\vec{V} = u\vec{N} \times u\vec{AB}$。三个单位正交向量 $u\vec{V}$，$u\vec{AB}$，$u\vec{N}$ 组成新坐标系的一组基。基于点 $A$ 构建新坐标系的四个基点：$A$，$u = A + u\vec{AB}$，$v = A + u\vec{V}$，$n = A + u\vec{N}$。这四个点依次对应于新坐标系中的 $(0,0,0)$，$(1,0,0)$，$(0,1,0)$，$(0,0,1)$。求解仿射变换矩阵：

$$M * \begin{bmatrix} A_x & u_x & v_x & n_x \\ A_y & u_y & v_y & n_y \\ A_z & u_z & v_z & n_z \\ 1 & 1 & 1 & 1 \end{bmatrix} = \begin{bmatrix} 0 & 1 & 0 & 0 \\ 0 & 0 & 1 & 0 \\ 0 & 0 & 0 & 1 \\ 1 & 1 & 1 & 1 \end{bmatrix} \quad (19)$$

式(19)得到的 $M$ 应用于所有平面上的点即可将坐标转化为 $(x, y, 0, 1)$ 的格式，取前两位 x 和 y 值完成三维到二维变换。

仿射变换后同样采用 RANSAC 算法在二维平面上拟合轮廓圆，拟合效果如所示。

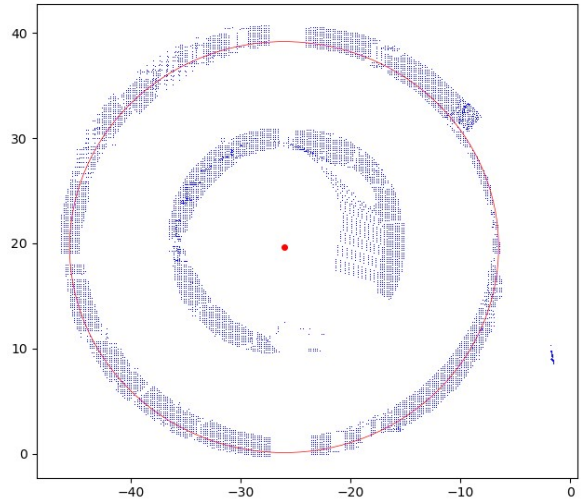

图 7 RANSAC 拟合得到端面圆

# 三、 实验结果与分析

## 3.1 实验设计

为了验证算法的有效性，在 V-REP 仿真环境中搭建测试场景，如图 8 所示。

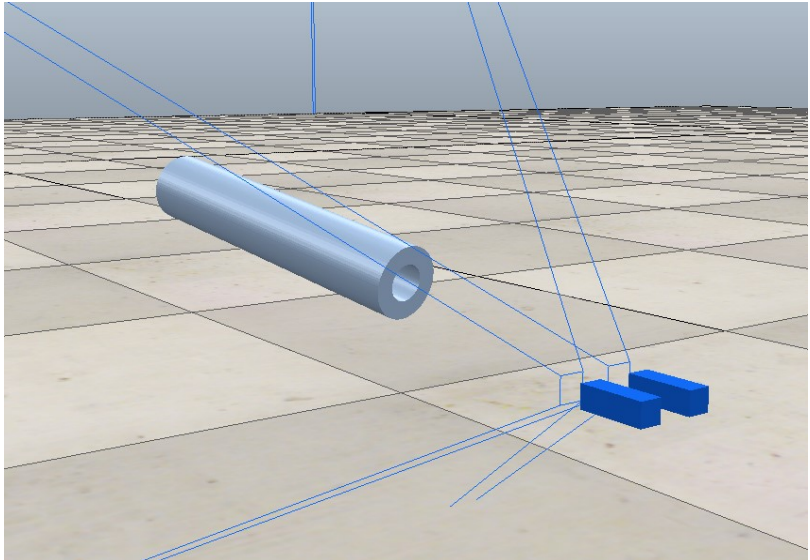

图 8 V-REP 中的测试场景

工件在实际环境中可能以任何姿态面向摄像头。设定套筒的两端面一侧为首一侧为尾，工件旋转 0~90 度时首端朝摄像头方向；旋转 90~270 度时首端被遮挡，尾端出现在镜头里；旋转 270~360 度时尾端被遮挡，首端再次出现在镜头里。由于首端和尾端相同，每个首尾端出现的 180 度里套筒的姿态自成对称关系，因此测试某个端面与摄像头 0~90 度夹角之间的旋转即可对应所有工况。

### 3.2 实验结果

测试用的工件长 200mm，内径 20mm，外径 40mm。定向误差定义为提取出的轴线向量与真实轴线向量的夹角，定心误差定义为提取出的端面圆心坐标与真实圆心坐标的距离。针对 0~90°之间的夹角，每隔 10°取一个采样角度，每个采样角度测试 14 次，记录 14 次数据的平均值，如表 1 所示。

表 1 不同夹角下的误差统计

| 夹角（度） | 定向误差（°） | 定心误差（mm） |
| --- | --- | --- |
| 0° | 0.88 | 1.97 |
| 10° | 1.08 | 3.39 |
| 20° | 0.91 | 1.95 |
| 30° | 0.61 | 2.67 |
| 40° | 0.98 | 3.20 |
| 50° | 0.78 | 3.64 |
| 60° | 1.17 | 4.14 |
| 70° | 0.79 | 5.13 |
| 80° | 端面被遮挡 | |
| 90° | 端面被遮挡 | |

如图 9 和图 10 所示，黑色射线的端点和方向为真实圆心和真实轴线向量，绿色平面为提取出的端平面，红色射线的交点为圆心的计算值，垂直于端平面的红色射线为轴线向量的计算值。

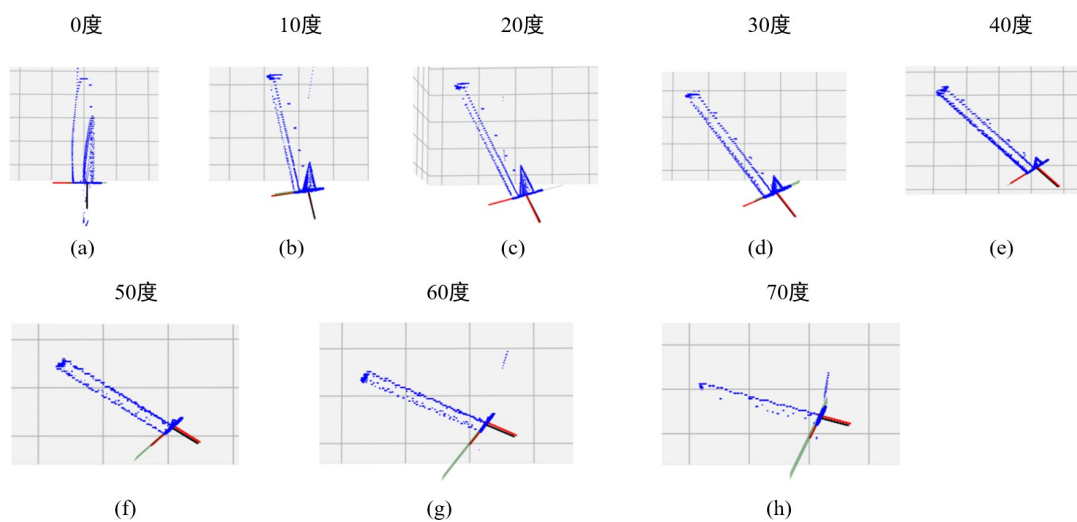

图 9 实验结果的俯视图

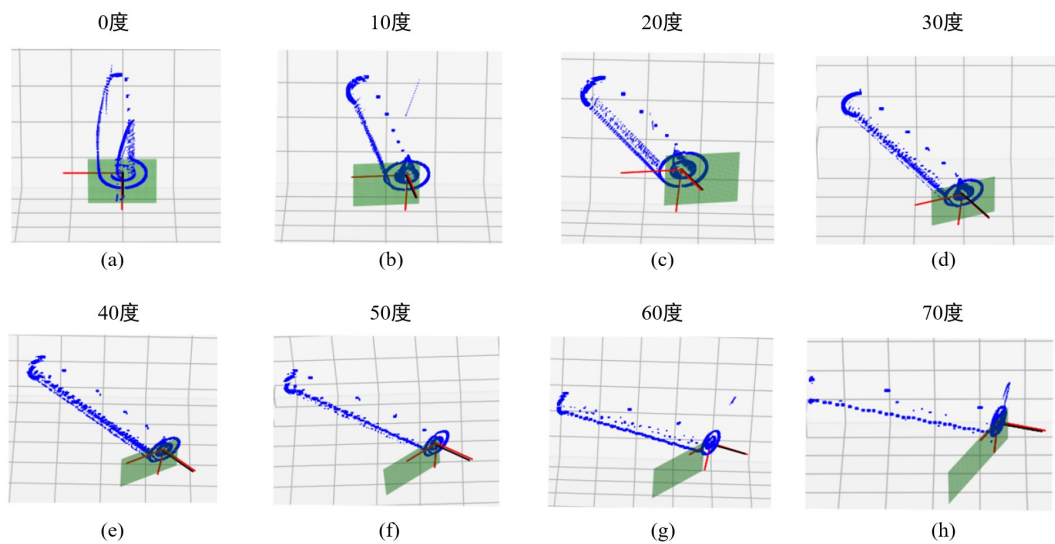

图 10 实验结果的侧视图

算法的实时性与精度同等重要。为了测试算法在不同角度下的运算时间，针对 0~90°之间的夹角，每隔 10°取一个采样角度，每个采样角度测试 14 次，记录 14 次数据的平均值，如表 2 所示。

表 2 不同夹角下的计算时长统计

| 夹角（度） | 定向时间（s） | 定心时间（s） |
| --- | --- | --- |
| 0° | 60.00 | 26.55 |
| 10° | 6.09 | 55.45 |
| 20° | 9.00 | 27.27 |
| 30° | 34.20 | 50.39 |
| 40° | 60 | 52.70 |
| 50° | 60 | 26.64 |
| 60° | 60 | 29.21 |
| 70° | 60 | 9.32 |
| 80° | 端面被遮挡 | |
| 90° | 端面被遮挡 | |

为便于比较，将结果绘制成图表，如图 11 所示。

随着夹角增大，定向误差的大小较稳定，无明显变化，定心误差则呈现逐渐增大的趋势。

定向误差来源于算法误差，误差大小受端平面点云厚度和点云密度影响。本实验中端平面能采集到足够多的点，即使在 70 度的夹角下端平面的点也能达到 1200~2000 个，满足高精度拟合平面需求，因此不同夹角下点云密度的影响不显著。点云厚度方面，随着夹角增大点云厚度会增大，但厚度的增幅较小，对 RANSAC 提取平面产生的干扰小，且 RANSAC 算法本身具有随机性，二者抵消使得点云厚度的影响不显著。因此，在不同的夹角下，方向误差稳定在 0.6~1.2 这一较小范围内。

定心误差来源于算法误差和三维重建误差。夹角较小时，定心误差主要为 RANSAC 算法误差；当夹角增大到 40 度以上，三维重建的点云逐渐远离真实位置，随着夹角增大偏离程度越大，此时的误差为算法误差与三维重建误差的叠加。

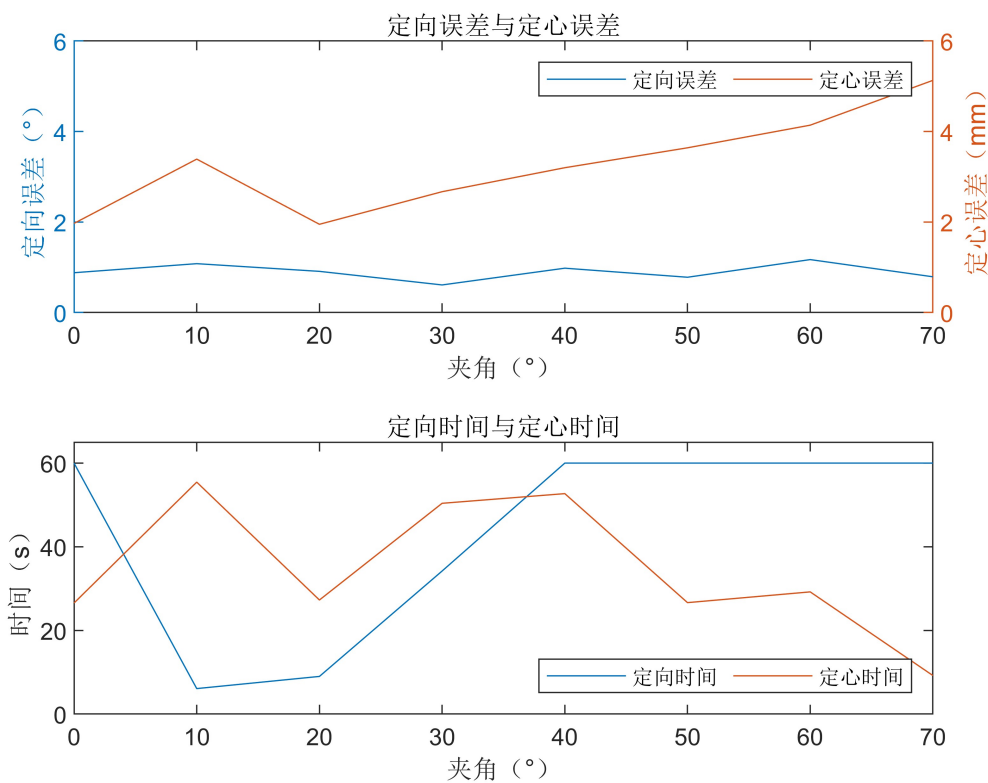

图 11 误差与计算时长

定心误差的算法误差与过拟合和点云畸变相关。端平面点云中点的数量最少也会达到 1000 个，这对 RANSAC 算法是较大的计算负担。实际计算时按一定比例抽样，然后根据样本计算圆心。若抽出样本过多，则容易产生过约束，这一现象体现在夹角等于 10°时，此时误差异常突增是由于在该夹角下最利于立体匹配，采集到最密集的点云，RANSAC 提取圆心时产生过约束，反而无法精确拟合真实圆心，使定心误差出现反常增长。

定向定心时间取决于点云的密度，密度大时拟合时间长，密度小时拟合时间短。

## 四、 结语

本文基于无人车载双目视觉系统，针对空间中任意姿态的圆柱形套筒开发了一种高精度的定位系统。运料车间中的无人车抓取工件时需要对工件进行精确定位，本文设计的系统可为无人车提供较准确的工件位姿信息。首先，分析比较各种方案优劣后，本文确定了平行双目视觉系统的布置方案；其次研究了双目相机的标定方法，在张正友标定法的基础上提出了一种结合传统标定法的改进方法；接着研究并比较了不同立体匹配算法，优选基于区域的立体匹配算法以及 SAD 灰度匹配相似性度量函数进行立体匹配；然后研究并比较不同的位姿估算方法，优选分步提取的工作流程，并选择 RANSAC 提取特征；最后设计实验验证了双目定位系统的有效性。实验显示，以表面无纹理、无明显特征的光滑圆柱体作为目标物体，在 0~90 度夹角下，系统的定向精度为 0.61~1.17mm，定心精度为 1.95~5.13mm，定向时间为 6.09~60s，定心时间为 9.32~55.45s，证实了系统的通用性、可靠性和精确性，能为运料无人车提供工件位姿信息。

高精度双目视觉系统还可用于识别小体积障碍物、悬空障碍物等传统手段难识别的潜在危害。在未来的工作中，作者将优化算法识别速度，引入新的数据预处理方法改善点云分辨率不足的问题，使算法迁移识别小体积、悬空障碍物。

# 参考文献